# Survey of Bayesian Network Applications to Intelligent Autonomous Vehicles (IAVs)

*Rocío Díaz de León Torres, Martín Molina, Pascual Campoy*

*Computer Vision and Aerial Robotics Group, Centre for Automation and Robotics (CAR) UPM-CSIC, Universidad Politécnica de Madrid, Calle José Gutiérrez Abascal 2, 28006 Madrid, Spain*

## Abstract

This article reviews the applications of Bayesian Networks to Intelligent Autonomous Vehicles (IAV) from the decision making point of view, which represents the final step for fully Autonomous Vehicles (currently under discussion). Until now, when it comes making high level decisions for Autonomous Vehicles (AVs), humans have the last word. Based on the works cited in this article and analysis done here, the modules of a general decision making framework and its variables are inferred. Many efforts have been made in the labs showing Bayesian Networks as a promising computer model for decision making. Further research should go into the direction of testing Bayesian Network models in real situations. In addition to the applications, Bayesian Network fundamentals are introduced as elements to consider when developing IAVs with the potential of making high level judgement calls.

## 1. Introduction

Autonomous Vehicles are becoming more popular nowadays. Ranging from ground, aerial, and maritime to even space vehicles. Their applications range from activities of day to day use such as transporting passengers to more critical activities including combat and exploration(underwater, space). Their use carries with it benefits and risks. They are capable of making decisions in situations in which humans could not do so, but we must ensure in some way that they make the right decisions so as not to put at risk the missions and the lives of people. The operational risk levels of Intelligent Autonomous Vehicles determine their acceptance by those who are intended to use their services. In order for an IAV to make a correct decision, it must consider several factors. These factors must be carefully selected into a context and represented, together with the relationships that exist between them.

Regardless of the environment in which IAVs move, they all face situations of uncertainty. There is a possibility that the sensors produce noisy measurements and there may be uncertainty in the performance of the autonomous vehicle. There is also uncertainty in the way in which humans behave (their performance) and in the behaviour of the other agents with whom they interact. The surroundings in which they operate are also uncertain, we can not have control of everything. The representation itself of the knowledge of the environment could be uncertain.

In the race for the creation of Autonomous Vehicles, Deep Learning has been the most popular computational model. It has been successfully used according to the Aerostack architecture **[Carrio et al., 2017]** for feature extraction, planning, situational awareness (perception), and motion control. However, if we consider the Aerostack architecture, we are missing applications to systems that receive high-level symbolic commands and generates detailed behaviour sequences (*executive system*). We are also missing applications to systems that supervise the performance of the Autonomous Vehicles (*supervision system*). According to **[Sünderhauf et al., 2018]** the application of Deep Learning in robotics motivates research questions: *How much trust can we put in the predictions of a Deep Learning system when misclassifications can have catastrophic consequences? How can we estimate the uncertainty in a deep network's predictions and how can we fuse these predictions with prior knowledge and other sensors in a probabilistic framework? How well does Deep Learning perform in realistic unconstrained scenarios where objects of unknown class and appearance are encountered?*

Decision making stays as an open question **[Gopalswamy and Rathinam, 2018]** because the autonomous vehicle do not only need to identify and classify the physical world, it needs to identify and classify the human intention (from the other drivers and pedestrians).

All these issues have made researchers consider Bayesian Networks as an alternative to let the Autonomous Vehicles deal with uncertainty and make intelligent decisions. Until now, the performance of the AVs depends to a high degree on human decision, however the current trend is to replace human decision making with algorithms. In **[Kochenderfer, 2015]** Bayesian Networks are presented as a suitable probabilistic computational model to deal with decision making under uncertainty. They are a good tool to model a variety of problems in different contexts. They are supported by well established algorithms to learn their parameters and build their structure. Researchers have developed robust algorithms to precisely track the reasoning process called inference in a Bayesian Network. In this way an Intelligent Autonomous Vehicle could be able to justify its decisions. Because of existing Bayesian Networks algorithms and their graphical essence to represent problems, the decisions can be analized quantitatively and qualitatively. This paper reviews BNs from two perspectives: (1) the kind of tasks related to Autonomous Vehicles that have been devised using Bayesian Network models (2) their foundations that need to be considered to develop Intelligent Autonomous Vehicles with the potential of making high level judgement calls.

In the next section, the general context in which an Autonomous Vehicle has to perform is described. Its interaction with the real world is considered as well. Here the operational factors of an IAV are presented. Section 3 covers BN concepts; learning, inference and verification algorithms and software. Afterwards, significant work on how operational factors have been modelled by BNs is discussed in Section 4. Cases of IAVs performing in ground, aerial, space, and maritime environments are included. BNs contributions to IAVs and challenges are stated in Section 5. Section 6 discusses open research, specifically in the areas of safety and collaborative work carried out by IAVs. Section 7 comes to conclusions about the present analysis.

## 2. Unmanned Intelligent Autonomous Vehicles

In this article the word Intelligent has been used to emphasize the idea of Autonomous Vehicles making high level decisions, for example, in complex situations a combat autonomous airplane has to decide attack, retreat or defend **[Cao et al., 2010]** during a possible failure in communications. Under this context the IAV has to make a decision by its own, however, there are other circumstances under which and IAV has to make decisions to collaborate and interact with humans or others IAVs.

Considering the current advances in technology is becoming more common we find IAVs in our daily lives. In some countries the government has granted permits to car and technological companies to test their vehicles on real conditions. We can see on the streets, testing cars in UK **[UKAutodrive, 2018]** and California state **[DMV California, 2018]**, **[macReports, 2018]**. An extensive area of the Trondheimsfjord in Northern Norway was designated as an official test bed for autonomous shipping **[Kongsberg, 2016]** by the Norwegian Coastal Authority (NCA) on September 2016. As one of the first coastal areas in the world officially dedicated to the development of technology for autonomous ships, the test bed is a vital facility for the future of shipping. In 2017 Norwegian government opened a third test bed in Horten **[Kongsberg, 2017]**, the new area is open to both Norwegian and international organisations, and is designed to be a convenient, safe, non-congested space to trial new technology and vessels. The European Authority in Aviation Safety (EASA) has review the Challenges and Opportunities of the Unmanned Aircraft System (UAS) **[EASA, 2017]**. They gathered in a conference senior aviation professionals from different backgrounds regulators, manufacturers, airlines and associations from all world regions to discuss global aviation safety issues from the perspective of both the regulators and industry. For spaceflights, numerous autonomous operations have been successful, however anomalies have occurred during shuttle operations and other demonstration missions. Fully autonomous spaceflights have thus far been limited, to extent them, there are proposals to include real-time autonomous decision-making and human-robotic cooperation. These topics are being investigated but have not yet been flight-tested **[Starek et al., 2016]**. The advancement of robotics and autonomous systems will be central to the transition of space missions from current ground-in-the-loop architectures to self-sustainable, independent systems, a key step necessary for outer-planet exploration and for overcoming the many difficulties of interplanetary travel.

In a real world humans and Autonomous Vehicles have to learn to coexist and interact within a legal framework, regulations, environment and social requirements. All this aspects are important to define the way they are going to share spaces and to operate individually and together. In **[Carrio et al., 2017]** they show Unmanned Aerial Vehicles components focusing on individual technical intrinsic factors and in **[Thieme and Utne, 2017]** they are presenting extrinsic factors so human and IAV can operate together. Intrinsic and extrinsic factors are going to influence the IAVs behaviour. By intrinsic factors the individual characteristics are considered, such as technical and reasoning skills. By extrinsic factors laws, organisational regulations, population requirements and nature restrictions are taken into account. With the inclusion of intrinsic and extrinsic factors is expected an IAV is able to reason about its own performance and be more aware of its surroundings. The *Figure 1* shows the operation framework and interaction between *humans* and IAVs, the decisions made by both humans and IAV will always be affected by intrinsic and extrinsic factors. We need a way to model human-IAV interactions, how humans and IAVs are reasoning, variables they are taking into account to come to a conclusion, and which ones are the origin of their decisions. The literature presents Bayesian Networks as a suitable approach to model IAV and humans behaviours, interactions including intrinsic and external factors. BN can provide graphical and numerical answers. We can see in what percentage the intrinsic and extrinsic factors affected the decision making, calculate the operational risk so we can make the relevant changes to ensure the mission success and safety.

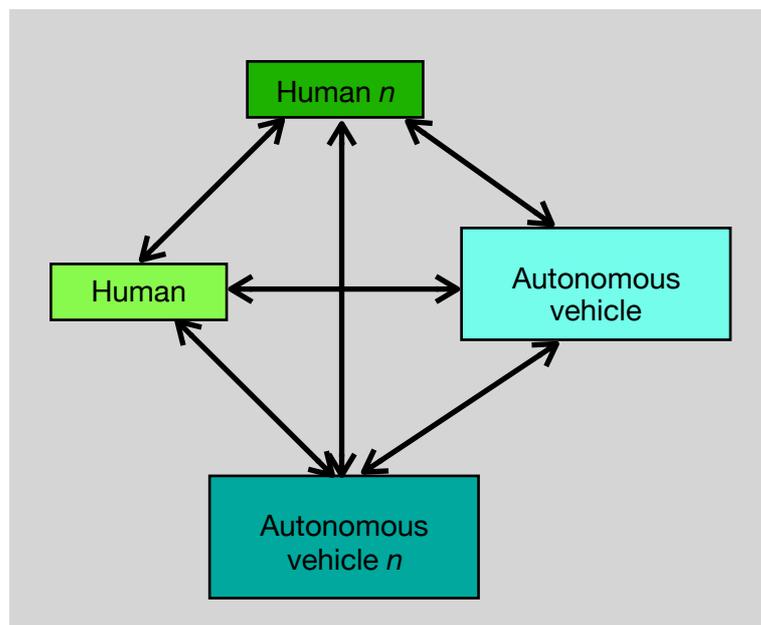

**Figure 1. Operation and interaction framework of IAV considering intrinsic and extrinsic factors.** Different colours are used to express humans and Autonomous Vehicles possess its own technical and reasoning skills, strengths and limitations (*intrinsic factors*). The grey rectangle means the relationships between humans and Autonomous Vehicles are also conducted under the shadow of environment, political and organisational regulations, social expectations and requirements, and economy (*extrinsic factors*). The links represent that the interaction between humans and Autonomous Vehicles could affect mutually their behaviour too.

# 3. Bayesian networks

Bayesian Networks are probabilistic graphical models used in Artificial Intelligence in a wide range of areas such as: speech recognition, medicine, stock markets, computer troubleshooting, text classification, computer vision, bioinformatics. As shown in the next section, they have been also employed in Intelligent Autonomous Vehicles. This section introduces BN properties and gives an outline of the algorithms and software to explore their creation and reasoning process.

Judea Pearl devised Bayesian networks in 1985 as a computational model for humans' inferential reasoning, namely, the mechanism by which people integrate data from multiple sources and generate a coherent interpretation of that data **[Pearl, 1985]**.

Causal networks, belief networks, probabilistic networks are other names for Bayesian networks. They let us represent knowledge based on conditional probability tables. Bayesian networks fundaments come from the Reverend Thomas Bayes' (1702-1761) theorem on conditional probability. Since the knowledge represented in the net is mostly *subjective*, *uncertain* and *incomplete*, it comes natural to interpret the data (reasoning) in them from probability theory. Network topology depends on data of the study area. It can be created from databases, expert knowledge, document or situation analysis.

Formally a Bayesian network (BN) is a directed acyclic graph in which the nodes represent propositions (or variables), the arcs signify the existence of direct (causal) dependencies between the linked propositions and the strengths of these dependencies are quantified by conditional probabilities **[Pearl, 1985]**. Each arc is directed from a parent to a child, so all nodes with connections to a given node constitute its set of parents. Each variable is associated with a value domain and a conditional probability distribution (CPD) on parent's values, *Figure 2*. The variables can take discrete or continuous values. Hence, it is possible to have hybrid BNs.

From machine learning point of view, BNs' learning is referred to construct a network automatically from direct observations avoiding human intervention in the knowledge acquisition process. For a deeper view on how BNs manage probabilities, their theory related to decision making under uncertainty and machine learning see **[Barber, 2012]**, **[Kochenderfer, 2015]**.

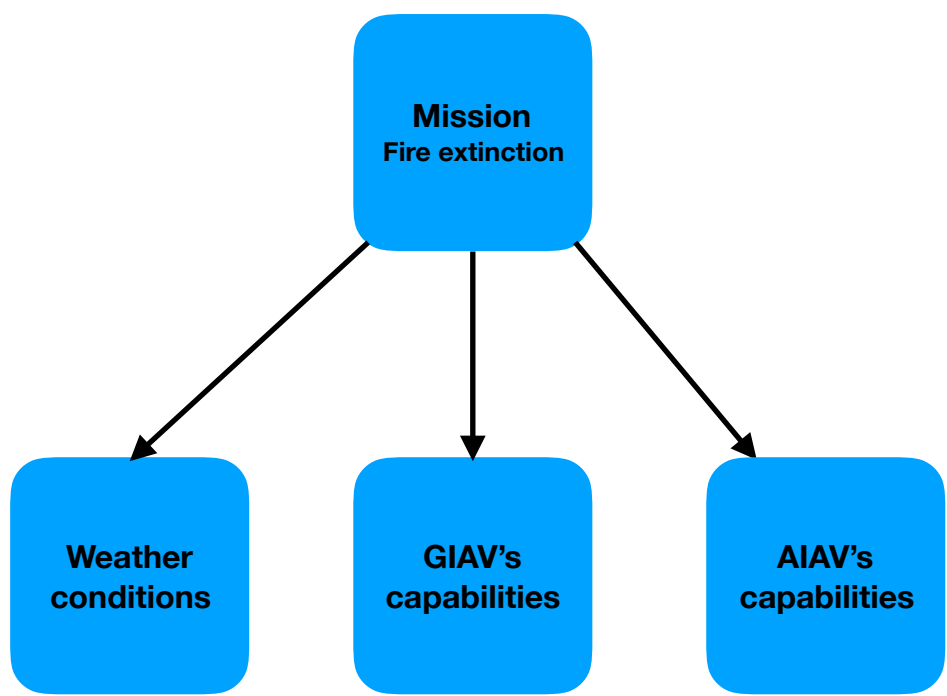

**Figure 2. Possible Bayesian network for forest fire extinction.** Two Intelligent Autonomous Vehicles could take part in the mission: ground (GIAV) and aerial (AIAV). The mission success of extinguishing the fire is related to weather conditions, GIAV's and AIAV's capabilities.

## 3.1 Algorithms for BNs

In order to exploit the BNs potential two main aspects have to be taken into account: learning and inference. The former has to do with defining the network topology (relevant variables, edges and edges directions) and network parameters (*a priori* probability distributions and conditional probabilities. The latter updates the probabilities of the variables in the network when new evidence is instantiated on specific values of some variables. For both, several algorithms have been developed, given good results under specific circumstances (network topology, continuos or discrete variables, exact or approximate solution).

*Learning* **[Neapolitan, 2004], [Gaméz et al, 2011]**. Design BNs in very complex domains, such as those random processes or situations from daily life, requires of methods to capture the structural relationship between variables from activities' data **[Doshi et al., 2002]**. The network structure can be created from expert's knowledge. Nevertheless, for the expert could be difficult to explain its knowledge and takes time to acquire it. Therefore, observations (data) have been considered to obtain and improve the network structure. In the learning of parameters, we start from the fact that we have the network structure and we obtain all the previous and conditional probability distributions from real data. In **[Heckerman, 1996]** several methods are presented to learn the structure and parameters of a BN, including techniques to learn with incomplete data; also addresses the supervised and unsupervised cases.

*Inference* **[Neapolitan, 1990**]. Once a network was built we can apply algorithms to manipulate the knowledge that represents. Inference (propagation) algorithms manage tasks such as:
-Belief updating: calculate a posteriori probability based on a BN (model, parameters and/or evidences) **[Jensen and Nielsen, 2007]**.
-Finding most probable explanation: the maximal posterior probability of a set of variables in the network given an observation of the values of another set of variables **[Kwisthout, 2011].**
-Finding maximum expected utility (MEU) decision **[Russell and Norvig, 2009]**.

## 3.2 Verification and validation

When representing knowledge in a BN of a specific domain that was extracted from experts and data, it is important to certify that the model constructed is acceptable for use. We require of methods to make sure the network was built according to the conceptual specifications (how well the network fits the data, *verification*) and to estimate its prediction accuracy (*validation*). **[Schietekat et al., 2016]** provide a methodology for validation and verification (V&V) of a Bayesian Network (BN) model for aircraft vulnerability against infrared missile threats. **[Ron, 1995]** review accuracy estimation methods and compare cross-validation and bootstrap methods. **[Friedman et al., 1999]** propose Bootstrap approach for confidence measures to induce better structures from the data and to detect the presence of latent variables. **[Zuk et al., 2006]** study BNs sample complexity to understand the number of samples needed in order to guarantee a successful structure learning. **[Van der Gaag et at., 2007]** present a survey on sensitivity analysis of probabilistic networks for investigating the robustness of the output of the net. The basic idea of the analysis is to systematically vary the assessments for the network's parameter probabilities over a plausible interval and study the effects on the output computed from the network.

## 3.3 Software

Many universities, research institutes and companies have developed software to work with BN. Some of the works presented in this survey have used the following:

*GeNie*. A tool for modelling and learning with Bayesian networks, dynamic Bayesian networks, and influence diagrams. It has been tested since 1998. Its compatible with other BN software. Includes learning and inference algorithms. There is a free academia version. Additional information can be see at: https://www.bayesfusion.com

*bnlearn*. An R package for learning the graphical structure of Bayesian networks, estimate their parameters and perform some useful inference. Since 2007, it has been under continuous development. For additional information see: http://www.bnlearn.com

Netica. A powerful, easy-to-use, complete program for working with BN and influence diagrams. It has an intuitive user interface for drawing the networks, and the relationships between variables may be entered as individual probabilities, in the form of equations, or learned from data files (which may be in ordinary tab-delimited form and have "missing data"). Under development since 1992. There is a free demo version that is full featured but limited in model size. For additional information see: https://www.norsys.com/netica.html

# 4. Applications

This section describes articles that have been selected to show different tasks related to IAVs. All the applications presented here involve the use of BN to model and analyse the problems. In some cases the BN configurations include just intrinsic or extrinsic operation-interaction factors and in others include both. We can see how real world variables have been considered to evaluate the performance of the autonomous systems, to represent the environment and infrastructure conditions. Specific situations on which the IAVs have to make decisions under uncertainty are also covered.

## 4.1 Ground vehicles

Nowadays, the recent autonomous car accidents have brought back questions such as: *How to make vehicles safer? How to detect cyber-physical threats in IAVs? Who or what is to blame?* The answers to this questions have been studied in the literature **[Li et al., 2016]**, **[Bezemskij et al., 2017]**, **[Gopalswamy and Rathinam, 2018]**.

The risk assessment for IAVs using Bayesian Networks has been approached in **[Li et al., 2016]**. They state a BN configuration as a framework to bring many factors together. The BN structure is created taking into account causal relationships of the variables, the root nodes are the evidences connected in a second level to event nodes (navigation, anti-attack, device state and control system), all of them are connected to a *central node use to detect when something wrong occurs to the IAV which do not meet the security requirement*. The central node is linked to a monitor node that protects IAV against failure, if a monitoring unit is not working gives an alarm. Because of the alarm must be received within a given period of time and with a given probability, the alarm monitor node is linked to a node representing if the alarm has been received within the specific time to alarm. For further analysis it is worth to mention that they have defined the network variables as binary. The main problem of IAV risk evaluation is how to determine what can be considered unsafe. This depends on the requirements of different standards of transportation. As a study case they present a simulation of a car equipped with GNSS/INS going through a city, the simulation time is 100s, however, there is a blockage of the GNSS navigation signals starting at 20s and ending at 60s. The IAV risk rises up when the navigation sources are reduced. Based on this approach the car designers could realise a sensitivity analysis in the aim to get a deep view of the IAV system and find the key factors of IAV security. The blockage of the GNSS navigation signals could cause the car loses its trajectory or has a erratic behaviour. Depending on what we include in this BN as event nodes an alarm can rise. For the simulation, problems in the navigation system activated an alarm. It can be notice when the car gets lost, but additional analysis is needed so it can consider possible alternatives to find its way and be able to think about consequences of its actions. Even though, they defined an anti-attack node, the way that protection measurements satisfy the requirements is not explored.

Like any computational system the IAVs are also subject to attacks. **[Bezemskij et al., 2017]** have studied the cyber attacks or sensory channel attacks (physical) which could affect the ability of an autonomous vehicle to navigate or complete a mission. They present a method based on Bayesian Networks that can tell when an autonomous vehicle is being under attack and whether the attack has been originated from the cyber or the physical domain. Detecting such threats is challenging, especially for resource-constrained systems, where highly accurate intrusion detection algorithms cannot be run on board and continuously. In a comparison with other methods proposed for threat detection such as Deep Neural Networks (DNN) for monitoring the CAN bus network and detecting malicious activity; in **[Bezemskij et al., 2017]**'s opinion DNN would be unlikely to be integrated into systems with resource constraint, since DNN are computationally heavy as they require a lot of processing power to teach the neurons using the data. Deep learning have shown being very accurate, but also an approach that consumes a lot of computational resources. If the Deep Learning Network is sufficiently reliable could be used offloaded to a more powerful infrastructure. This can reduce detection latency and reduce energy consumption, however, the major drawback is that it depends on the availability of an offloading infrastructure, which is impractical in many areas of Autonomous Vehicles. Nevertheless, we need to bare in mind the number of variables in a BN and wether they are discrete or continuous. A big network will consume a big amount of computational resources. In **[Bezemskij et al., 2017]** BN are presented as an approach that can be applied across a variety of vehicles and can adapt by learning what is normal for that vehicle and *detecting deviations*, so that is also applicable to unknown/probable threats. They demonstrate the validity of their approach on a ground autonomous robotic vehicle in a routine mission scenario, where it has a target which is to reach a destination, with *stochastic* elements that divert it. To detect the cyber-physical threats seventeen variables were defined representing the monitored features: battery voltage, compass bearing, pitch, roll, front distance, back distance, left distance, right distance, temperature, four variables each one for a motor, packet arrival time, action indicator, sequence number, CAN packet rate. Additionally, three variables to supervise the threat status were specified: cyber, normal (no threat) and physical. The data fed into the BN to train and query is the output of an heuristic binary classifier, which filters incoming data samples violating frequency of occurrences of the data samples within the deviation region. It learns the signature characteristics from the sensor data being transmitted to the CAN bus. The heuristic binary classifier achieves reasonably high threat detection rates. However, it does not have the ability to tell anything more about the source of the threat beyond its existence. To address this, they implemented a BN working with discrete data that can be in a qualitative or quantitative form. Besides, BNs are able to infer the unknown variables which are useful in a situation where an intrusion detection mechanism has to make a decision. To built, learn the parameters and query the BN they have used the open library *bnlearn* **[Scutari, 2010]**, which runs under the statistical analysis environment *R*. From *bnlearn* the *Hill-Climbing* algorithm **[Gámez et al., 2011]** was selected to built the network. In order to built a BN and learn its parameters training datasets are needed. Training datasets were created based on the three events to be queried given the evidence: *normal*, *cyber attack* and physical attack. 70 dataset were used for training and 30 data set for testing. In this work, it is important to notice that we need to count with relevant and enough data to make a good characterisation of the variables under study.

The previous two works deal with fails detection and source identification from where the sensor data alterations were originated. But who is about to blame when something goes wrong. In a real world we can not just consider the intrinsic operation factors of the IAVs, the IAVs can be also affected by the operational external factors. We have to think about vehicles interacting with different environments and people or even other vehicles. The current autonomous cars manufacturers shoulder the primary responsibility and liability associated with replacing *human perception* and *decision making* with automation. Until now, in a car that count with an autopilot when is about to collide, human response is required to take control and avoid crash. Nevertheless, when using autopilot humans become distracted and their reaction has not been fast enough causing accidents. A total automation of driving meaning *automating human decision making* could improve safety on the roads. That is the next big step that is still under discussion. **[Gopalswamy and Rathinam, 2018]** proposed an approach to apportion the responsibility and liabilities associated with autonomous cars between traditional automotive manufacturers, infrastructure players and third-party players. For them, the automotive original equipment manufacturers should concentrate on providing all the physical components that are needed for a car to move and perceive its surroundings, the manufacturers could also provide advance driver assistance systems. The analysis of the performance of the physical components should be separated from the analysis of the embedded software functionality. The software on board for autonomous cars would have the big responsibility to automate decisions to drive safely. High level decision making for cars it is more than just perceiving the world, means identifying and classifying human intentions (of other drivers and pedestrians), understanding the cars surroundings (objects near by, their state of motion, traffic signals, weather and road conditions), caring for its own safety and safety of the other ones. The car would move by driving by wired capabilities enabled by the manufacturer; the directions of displacement or car actions would depend on sensor data from the car itself and from the infrastructure operators. **[Gopalswamy and Rathinam, 2018]** present a concept of special infrastructure traffic corridors provided with road site units fitted with special devices to generate information about the situation on the road. A special device residing in the car will interface with the driving by wire capabilities to receive the traffic corridor information and provide the commands to drive autonomously, so the special device will make a decision and will instruct the vehicle to perform maneuvers. As we can see, different parties are involved in the complete autonomous car operation. Often multiple components fail simultaneously and in such cases, we need a mechanism to define the blame for the individual components. We can have different fault configurations that could result in the failure. To find the probability of the fault configurations they devised a general BN. The root nodes represent random variables associated to each of the components of the autonomous vehicle system. These root binary nodes denote if one or more of the components are at fault or not. They assumed that fault variables are mutually independent. An intermediate node represent the set of all the possible outcomes of the functioning of the system. The outcomes are determined based on their severity and their risk level. It is also assumed that the set of outcomes is mutually exclusive. There is a third level at the bottom, which contains discrete random variables, each one being associated to the cost of the faults of each component. So, the BN has links from the fault nodes to the outcome node and to each one of the cost nodes. The outcome node has links to cost nodes. The cost can be viewed as the "blame" or "responsibility" assigned to the component and depends on the fault configuration and the outcome of the system. The BN provides a model of the joint distribution of all the random variables in the system and their conditional dependencies. They infer the likelihood of an outcome for a given fault configuration of the components based on hazard analysis **[Beckers et al., 2013].** The paradigm presented by **[Gopalswamy and Rathinam, 2018]** establishes a clear way to separate and distribute responsibilities of the parties involved in the design, construction and performance of the autonomous cars. In the one hand, there is an emphasis in the split between the car manufacturers and the intelligent software developers for autonomous car decision making. On the other hand, new digitized roads are conceived. Even though a BN is defined to show the way the parties would be interrelated and how we could calculate the degree of responsibility, they just present an hypothetical case for its evaluation. However, the probabilistic nature of BN could also helps us in the analysis of more uncertain situations, for example, the performance of all-terrain cars. The real world is full of unexpected things and the BN are a good way to deal with the changes that an all-terrain car could find on its path.

## 4.2 Space and Aerial Vehicles

Modern aircrafts, whether for commercial, exploration or combat purposes, require special systems to operate in the absence of humans or to support pilots in their performance to make the right decisions. In order to operate autonomously, these aircrafts should be able to monitor the robustness of the on board software and sensor systems (intrinsic factors). The perception and analysis of the situations in which are immersed are also important for their autonomy (extrinsic factors). The air companies are investing on Artificial Intelligence to develop full autonomous airplanes to make flying even safer by removing the risk of human error. But still, there is a lot of to do when it comes to control the airplane and different situations during the entire flight. Experts consider humans still beat computers when it comes to deal with a medical emergency, security incidents, or a quick mechanical fix. There are several articles in the literature that show how the unmanned aerial vehicles could make decisions in complex environments replacing humans from the Bayesian networks perspective. Even that the different Bayesian networks approaches for IAVs have not been tested in real scenarios the lab experiments indicate they could work for high level decision making. People interacting with IAVs will be able to see how the decision are made based on the graphical model and inference algorithms of the BN.

In 2016 a book chapter **[Starek et al., 2016]** was published about spacecraft autonomy challenges for future space missions, they point out *real-time autonomous decision-making and human-robotic cooperation* as part of the biggest challenges. The spacecrafts automation has been studied for several years. One example in which an autonomous spacecraft has to select a landing site is formulated in **[Serrano, 2006]**. A landed exploration mission can be successful if a convenient final landing site is chosen. Three main variables are defined on which to select the landing site: *terrain safety*, *engineering factors* (spacecraft's descendent trajectory, velocity and available fuel), and *pre-selected by scientist multiple potential sites*. These variables are parents of a discrete random node representing landing quality in a Bayesian network. Nodes representing *terrain features* are children of the node *terrain safety*. The terrain features are a combination of discrete and continuous random variables. The craters and rocks are discrete and the slope and roughness are continuous. The network was built to model probabilistically the relationships between the variables that impact the quality of a landing site. To evaluate the proposed approach, experiments were performed using a dynamics and spacecraft simulator for entry, descent and landing. Also, digital elevation maps representing a variety of planetary terrains were created. The results indicate that the proposed approach selects the site that best optimises all three key variables. For instance, in one of the experiments there are two regions of high scientific interest. However, one of the regions lies near the boundary of the engineering factor area and the other lies in an unsafe part of the terrain. Consequently, a site with medium scientific interest is selected because it has a better balance of all three key variables. We can consider each one of the three main variables, being decision nodes by them selves. The *landing quality* node is a node that makes decisions when *integrating the information* of the main variables. The most relevant information for decision making should be selected in advance if we want the autonomous system to react in real time. For example, determining areas of high scientific potential is a laborious process that involves numerous considerations beyond the scope of an on board reasoning system. To influence the on board site selection, the scientists' preferred site can be designated in advance and integrated via the network. Depending on the characteristics of the mission, the variables and the relationships between them could change. Therefore, we would require the use of networks with variable structure.

Since relatively few years the interest in the decision making of Autonomous Vehicles has been growing. In **[Cao et al., 2010]** the situation assessment of an UCAV (Unmanned Combat Aerial Vehicle) is explored. Because of no pilot, the casualties vanish. But, the signal between the operator and UCAV may be disturbed and it may be miss the attack chance in the complex battlefield environment at present, UCAV strengthens the capacity of the autonomous attack and develops the autonomous decision system. Considering the concept of *situation assessment* which has been defined as the *perception*, *comprehension* and *prediction* to the elements in the environment within a volume of time and space; they modelled the decision making process based on dynamic Bayesian networks (DBN). In the decision making process is important to study variables that can give information about the possible reaction of the adversary and ensure the mission success. The opposite main situations are attack, defence and retreat. The information of discrete nodes is from the opposite intention and activity. To assess the opposite situation, decision maker usually acquires the more information from the sensors and airborne warning and control system. UCAV has to deal with a huge load of information and to be selective for relevant information. The collected information comes from different platforms and it is difficult for decision maker to deal with so much different information in limited time and under pressure. UCAV can take the autonomous decision making to plan and replan trails/task online, establish goals and deliver weapons according to the development of the battle field situation. **[Cao et al., 2010]** modelled a DBN as a tool for the UCAV to assess the situation. The DBN allows the UCAV to *fuse* a variety of perceptions (sensors, opinions, databases, situations) and finally gain the opposite situation. In the paper they used junction trees algorithm for inference. Since the algorithm is a static BN inference algorithm, they change the DBN to a static BN. The network structured and conditional probability table were defined based on experts' opinions, experiences and data learning through the computer. Initially with no information about the situation, it is assumed that the prior probabilities of nodes in DBN obey uniform distribution. After initialisation, the situation assessment system waits for the refresh data. Once new information is obtained, the reasoning of network is started up. The probability distribution of node state in network is refreshed, and the probability distribution of root node is obtained. With the change of time, the DBN can adopt reasoning according to the history and latest evidence information and its proved to have the ability to dynamically modify and improve the knowledge base and model of the situation assessment. Even that the simulation results proved to be similar to experts' results, the authors point out that military knowledge and experts' experiences are the key point for DBN. The structure and conditional probabilities of the networks depend on ways of thinking, experiences, multi source data and quantitative situation information for achieving autonomous decision making systems. Correct and rapid situation assessment has important research value and actual combat significance on tactical flight path, mission planning and distribution tasks.

Following with combat situations, in a real battle an IAV could face pop up threats, such as aircrafts/helicopters or ground based defence systems. The information about these threats could be incomplete (only partially available or not fully known). **[Gao et al., 2011]** have addressed the problem of online dynamic Unmanned Aerial Vehicle (UAV) path planning. To attack the problem they integrated a structure varied discrete dynamic Bayesian network (SVDDBN) into a model prediction control (MPC) algorithm. The SVDDBN is used to construct an online dynamic threat assessment model by estimating and predicting the states of the pop up threats, and the output of this model is then fed into a MPC algorithm for path planning. The network proposed allows to consider uncertainty about time, position and behaviour of the threats. The network is capable of inferring when some data is missing or incomplete, such as in the situation where a pop up threat randomly appears or disappears. A converted measurement Kalman filter (CMKF) algorithm is integrated into the Bayesian Network to allow estimation of the missing data about the threats. The mission of the UAV considered in this paper is to fly through, as safely as possible, a high threat area where a manoeuvring pop up threat could exist. Attention is confined to a two dimensional (2D) representation of the situation. It is assumed that at any given time during the flight the high threat area is a circular area with a pop up threat at the centre of the circle. A set of associated, dynamically changeable threat circles exists when the pop up threat changes in nature. All the nodes in the network represent information that is updated at every time slice *t*. The root nodes represent the threat probability, and the intermediate and bottom nodes represent threat information such as probability of UAV finding a threat if it flies along the movement direction; distance between the UAV and the threat centre; completeness of information; indication of whether the distance between the UAV and a threat has changed, i.e. if the distance has increased, decreased, or remains unchanged; radius of the threat circle; angle between the movement directions of the threat and UAV and estimated threat state (i.e. missing data), which mainly involves the position, speed, acceleration and the movement direction of the threat. The threat probability was predicted using the threat assessment network and the forward inference algorithm. The results of the simulation showed that the selected parameters allowed in the tests to generate the required position in a shorter time to the flight time, and this satisfies the requirement for generating UAV control in real time. However, special attention should be paid to select parameters that can let IAV make decision in real time without compromising to make correct decisions.

Fly-by-wire commercial aircrafts and UAV are fully controlled by software. Failures in the software or a problematic software-hardware interaction can have disastrous consequences. This motivated **[Schumann et al., 2011]** work. They created a Bayesian network as the modelling and reasoning paradigm to achieve an onboard software health management (SWHM) system. A SWHM system monitors the behaviour of the software and interacting hardware during system operation. In addition, a SWHM system performs diagnostic reasoning in order to identify the most likely root cause(s) for the faults(s). This is particularly important, since many software problems do not directly manifest themselves but rather exhibit emergent behaviour. For UAVs, the available bandwidth for telemetry is severely limited; a dump of the system state and analysis by the ground crew in case of a problem is not possible. For manned aircraft, a SWHM system can reduce the pilot's workload substantially. With a traditional on board diagnostic system, the pilot can get swamped by diagnostic errors and warnings coming from many different subsystems. Because of many software problems occur due to problematic SW/HW interactions, both software and hardware must be monitored in an integrated fashion. The network includes relevant nodes for file system related faults such as Status *File_System*, *Sensor File System Error*, *Delay*, *Health File System*, *Write File System*, *Sensor Queue-length*, *Status Message_queue*, *Sensor Oscillation* to represent software and hardware elements situations. The links between the variables specifies the interaction between them. To simplify design and execution, and at the same time be able to manage temporal aspects (without having to use a dynamic Bayesian network), preprocessing steps were carried out. For demonstration purposes, they designed a flawed software architecture with a global message queue that buffers all control signals and logs them in the file system before forwarding them. The following scenario was run: the file system is initially set to almost full. Subsequent control messages, which are being logged, might stay longer in the message queue, because the blocking write into an almost full file system takes substantial time. This situation potentially causes overflow of the message queue or leads to loss of messages. However, even if a small delay (i.e., a control message is not processed within its allotted time frame, but one or more time frames later) can cause oscillation of the entire craft. This oscillation, similar to PIO (pilot induced oscillation) can lead to dangerous situations or even loss of the aircraft. In this scenario, the software problem does not manifest itself within the software system (i.e., in form of errors or exceptions). Rather, the overall behaviour of the aircraft is impacted in a non-obvious way. The SWHM reasoner can then disambiguate the diagnosis by evaluating whether the fault is due to PIO or a software problem. The Bayesian networks for SWHM systems can be used to model software as well as interfacing hardware sensors, and fuse information from different layers of the hardware-software stack. **[Schumann et al., 2011]** concluded that Bayesian networks system health models, compiled to arithmetic circuits, are suitable for on-board execution in an embedded software environment. However open questions remain, How to handle unexpected and unmodeled failures? How to more automatically generate SWHM Bayesian models based on information in artifacts? They suggest to include software engineering models, source code, as well as configuration and log files. A good software-hardware interaction can take us to an error free performance of the UAV from the operational intrinsic factors point of view. The IAV could be able to achieve the tasks for which was made. Nevertheless, an IAV would need an additional element to reason about if would be useful to performed the task in certain time under specific situations. In other words, operational external factors could be considered towards IAVs.

Either for combat or rescue issues, for example, the way to achieve cooperation between Autonomous Vehicles has been researched. In particular, the case of searching by UAVs was analysed by **[Guo et al., 2012]**. They modelled the capabilities of heterogenous UAVs with a tree shape Bayesian network for each one, the network structure was derived from expert knowledge. The root node represents sensing capability and bottom nodes represent the detection width, flight velocity. We can include more UAVs' attributes adding more nodes. To verify the effectiveness of their searching area decomposition approach, simulation experiments were carried out in Matlab 7.1 environment on a laptop with 1.6GH CPU processor and 512M memory. In this simulation, three UAVs have to search an area defined by a convex polygon with eight edges. First the UAVs' relative capability probabilities are computed reasoning with the Bayesian network for each UAV. Later, each UAV is assigned an area (convex polygon) according with its relative capabilities. Then, each UAV plan the waypoints needed to follow a zigzag pattern to those directions. Distance between parallel lines depends on the sensing width of the UAV. If any of the initial parameters to calculate the UAV's sensing capability is missing temporarily, say an angle that determines the field of view, the area decomposition task can not be taken anymore by any existing deterministic decomposition methods. Thanks to the Bayesian network approach, still it is possible to infer by a priori knowledge in this uncertain situation. Therefore the Bayesian network approach has advantages over other methods on coping with uncertainty. **[Guo et al., 2012]** presented a model to evaluate the individual skills. The global performance of a team is also an important factor to estimate in the future, in order to warrantee the mission success, as well as, to extend the model to the cooperation of various Autonomous Vehicles aerial, ground and maritime.

When we have a scenario where many parties are intervening and many events are taking place, such as in a battle field; a method to assess the damage is required to review the relationships and influences between the parties and the effect of the actions to be taken. [**Chen-han and Jian, 2014**] built a Bayesian network for battle damage assessment. In the modelling process the nodes and parameters were gotten from assessing system based on professionals experience, information sources and their relations, the internal effects on information sources. After the network was modelled, they apply inference to assess the damage of UAV to ground attacking, testing on hypothetical cases, in the attacking between UAV formation and airfield runway from enemies. Discrete values are defined for the node for real damage assessment: no damage, light damage, moderate damage, severe damage and completely destroyed. The variables were define according to the following factors: environmental differences of the objects, the characteristics varieties of the targets, direct relations between attacking effects and targets, information of the data base and the professionals' experience. In **[Chen-han and Jian, 2014]** for the study of the network, nodes and parameters were simplify. In a real world scenario, we have to select the most relevant variables for the decision making. A big size network would take us to delays to carry out actions. However, Bayesian networks are a convenient tool to help a commander to have graphically an overview of the problem and to simulate many situations, in order to foresee attack actions. In this case, networks of variable structure would be needed. In some point, in the inference process the interaction between the network estimations and expert criteria should be taken into account.

If we want people trust and accept Intelligent Autonomous Vehicles, we should ensure they are capable of evaluating situations in which humans are involved and taking the best choice. Unmanned Aerial Vehicles (UAV) must follow governments' rules for people safety, to be allow into the civilian airspace. UAV must be trained to handle emergency situations as a pilot usually would do it aboard a manned aircraft. In the specific emergency case of a forced landing, UAV has to decide the most suitable forced landing site from a list of known landing sites. A solution for this critical situation was formulated by **[Coombes et al., 2016]**. Based on the specifications for a forced landing system laid out in a NASA Technical report (Civil UAV capability assessment), three main criteria were developed for selecting a suitable location for an UAV to attempt a forced landing in order of importance: risk to civilian population, reachability and probability of a safe landing. The emphasis is on public safety, where human life and property are more important than the UAV airframe and payload. Mitigating risk to civilian population has a much higher weighting than site reachability and safe landing since aircraft survival is of a lower priority. General Aviation (GA) pilots consider technical factors in making a decision for UAV safe forced landing. These factors are considered to estimate the probability of a safe landing. Factors taken into account by pilots are: wind, obstacles, size, shape, slope, surface. All the previous benchmarks, were included in a Multi Criteria Decision Making (MCDM) Bayesian network. The root node is a decision node to select the site for landing. Each discrete state of the root node represents each possible site, the number of states is determined by the number of landing sites. The network was tested by simulation, using a pre-mapped area with fourteen known field locations and discrete parameters. The scenario was a Cessna 182 in climb-out after taking off from Nottingham aerodrome. It had an engine failure at 400 meters above ground level with a wind speed of 10m/s from 270º. According to the network landing site 13 was the site chosen to attempt the forced landing into, as it has the highest marginal posterior probability. It was the favoured choice because it was a long field with over 50% extra length required, medium reachability, a safe grassy surface, free from obstacles, far from the civilian population. They solved the Bayesian network using diagnostic reasoning to improve computational speed and enable real time decision making. Pre-analyses were taken into account to structure the network in a different way. **[Coombes et al., 2016]** model gives an overview of *multi criteria decision making*. New types of nodes are included, *decision node*, *utility node* and *criteria nodes*. The criteria nodes have a direct influence on the utility node. The selection of the field is based on the field that gives the highest utility. To apply this model in real life has to be tested covering a larger number of cases, calculated the risks and consequences of the decisions.

The current trend is human-IAVs coexistence. In the future we will get mixed working groups supervised by humans or autonomous systems. In the human working groups, the responsibilities can be delegated and individuals can autonomously take actions according to workload increases. Accordingly a group of drones becomes bigger, mental-workload for monitoring them will be more demanding for a human supervisor. Flexible levels of autonomy could reduce supervisor work. **[Bazzano et al., 2017]** predicted the level of autonomy via a Bayesian network classifier in drone-traffic control tasks. In the bottom node three *levels of autonomy* are proposed: warning, suggestion and autonomous. The system warns the operator if critical situations occur, suggests feasible actions to him or *monitors and performs actions autonomously* without any human intervention. A node *mission outcome* has a direct influence on level of autonomy, and is a child of *workload* node. It was considered that the probability of changes in operator workload is conditioned on changes in the number of drones in three *alert nodes*: safe, warning and danger. Initial experiments were carried out with people to collect data to train the classifier to learn how to determine the appropriate level of autonomy for the system. A cross validation technique (training data vs testing data) was used to test the classification model performance. The Bayesian network training phase was performed by the Netica software, then the validation methodology was performed by obtaining a classification level of autonomy equal to 83.44%. The NASA-TLX questionnaire was taken into consideration as subjective workload assessment technique. More variables that can give a wider view of humans performance need to be included, such as physical and mental skills.

## 4.3 Maritime Vehicles

All kind of vehicles are going into the direction of becoming autonomous and Maritime Vehicles are not the exception. The environment in which IAVs have to perform involves different risks and variables, wether are moving on ground, air, space or water. All of them involve movement restrictions and acting conditions. Several studies have been carry out to evaluate the scenarios that maritime vehicles have to face. This section shows some examples about selection and evaluation of the variables represented in a Bayesian network model. Two articles have been chosen. The first one is about *monitoring the behaviour* of maritime traffic **[Lane et al., 2010]**. The second one evaluates the risk of interaction between humans and autonomous underwater vehicles **[Thieme and Utne, 2017]**.

According to IMO (International Maritime Organization) international shipping transports more than 80 per cent of global trade to peoples and communities all over the world. Shipping is the most efficient and cost-effective method of international transportation for most goods; it provides a dependable, low-cost means of transporting goods globally, facilitating commerce and helping to create prosperity among nations and peoples. The world relies on a safe, secure and efficient international shipping industry. All the traffic in the seas needs to be monitored. The transport of goods and passengers has to be safe and the marine and atmospheric pollution has to be avoided. Ships involved in commercial activities tend to follow set patterns of behaviour depending on the business in which they are engaged. If a ship exhibits anomalous behaviour, this could indicate it is being used for illicit activities. The IMO's International Convention for the Safety of Life at Sea requires automatic identification system (AIS) to be fitted aboard international voyaging ships with 300 or more gross tonnage (GT), and all passenger ships regardless of size. Ships with AIS transponders transmit their location, course, speed, and other details, such as their destination and ship identifier, at regular intervals. The analysis of this information to *identify behaviour patterns* has been studied by **[Lane et al., 2010]**. An overall threat is often manifested by a series of individual behaviours. An example threat scenario is the illegal exchange of goods at sea. The behaviours exhibited by a ship undertaking this activity could include deviation from standard route, turning off an AIS transmitter, entering a zone known for illegal exchanges and close approach with another ship. Thus, five abnormal ship behaviours were outlined to be presented in AIS data: deviation from standard routes, unexpected AIS activity (transmitter switched off or false position has been given), unexpected port arrival, close approach and zone entry. For each behaviour, a process is described for determining the probability that is anomalous. Individual *probabilities are combined using a Bayesian network to calculate the overall probability that a specific threat is presented*. The root node (*threat type*) carry out the anomaly *fusion* for threat assessment. The middle nodes (*behaviour variables*) are root node's children. Each of this variables takes a binary value, indicating that the behaviour is present or not. The nodes representing the *observed detector outputs* are children of *behaviour variables*. So the *threat assessment* is influenced by the *detector outputs* via *behaviour variables*. A numerical example is given to illustrate the principle of the Bayesian network for threat assessment. The results depend on interpreting outputs of anomaly detectors as probabilities.

Deep water exploration comprises a big risk for humans. Nowadays, autonomous underwater vehicles scan the seabed while being monitored by humans. Since underwater vehicles are very expensive and difficult to recover, a fail in an exploratory mission could result in big loses. Hence, *risk models are needed to assess the mission success forehand* and adapt the mission plan if necessary. *A general scenario where humans and autonomous underwater vehicle collaborate has been represented in a Bayesian network* by **[Thieme and Utne, 2017]**. Based on a case study they came into the conclusion that the human-autonomy vehicle collaboration can be improved in two ways: (1) through better training and inclusion of experienced operators and (2) through improved reliability of autonomous functions and situation awareness of vehicles. Their human-autonomous collaboration Bayesian network can improve autonomous underwater vehicle design and autonomous underwater vehicle operations by clarifying relationships between technical, human and organizational factors and their influence on mission risk. **[Thieme and Utne, 2017]** follow a well structured five steps process to develop the Bayesian network:

1. Describe aim and context of the BN

    The aim of the model in the article is to show the relationship between human operator performance and the technical performance of the autonomous system. The aim of the model determines the definition of the root node called *HAC* (*human-autonomy collaboration*) *performance*, which can take one of two values (inadequate, adequate).

2. Gather and group information relevant for the context into nodes

    The literature was used to group the information into two overall categories: (1) *autonomous function performance* and (2) *human operator performance*. The performance of the two main agents is monitored through these nodes. Besides, the previous nodes a third node called *level of autonomy* have a direct influence on *HAC* node. A total of 24 nodes were define.

3. Connect the nodes with directional arcs

    The links between variables (arcs) in the BN were defined based on the findings in the literature and the relationships identified between factors. It was found out that some factors have a mutual influence on each other. However, since BN are acyclic, in this article the most frequently mentioned direction of influence defined arcs.

4. Determine the conditional probability tables and quantify the model

    The data for the input nodes in the model were derived from a case study, (with basis in autonomous underwater vehicle operation) in the Autonomous Underwater Robotics Lab at the Norwegian University of Science and Technology.

5. Test and validate the model

Since they present a new situation for which there are not enough historical data to contrast the results obtained by BN, the BN model was evaluated considering that was built based on the literature and it became structurally similar to previous models.

Many factors were considered for building the net. However one stands out: *Trust*. *Trust* is a special factor in human-autonomous vehicle interaction. As in any relationship between humans, *trust* in the autonomous vehicle is built with time the human operator has been interacting with it. *Trust* also depends on the Operators' Experience, Operators' Training, Feedback of the Vehicle and Reliability of Autonomous Functions. Workload, Operators' Experience, Time Delay of Transmission and Operators' Training influence the Reaction Time of Operators. All these are examples of the factors that were included in the net. GeNIe 2.0 was used to conduct a sensitivity analysis, varying each node over the whole range and assessing the impact of this change on the target node. The target node for the sensibility analysis was the *Human-Autonomy Performance Collaboration* node. In the results of this analysis is noticed that the most influential input nodes on the HAC node are Autonomous Function Performance, Reliability of Autonomous Functions, Situational Awareness of Vehicles, Operators'Training and Operator's experience. This is an article that represents a whole view of intrinsic and extrinsic factors integration. The model needs to be evaluated with real data and work with networks that can affect the human and autonomous vehicle performance, for instance, networks for: environmental interactions, technical system performance, societal expectations and regulatory and customer requirements. Additional analysis is needed for variables that present mutual influences to determined how they affect to each other performance. The validity of the model is assumed, but hasn't been quantitative verified.

## 4.4 Analysis

The purposes of this article are to find out the works that have been developed using Bayesian networks for IAVs applications, to know how they have been used and their results. After the search, the estate of the art is pointing out in the direction of *high level decision making* as a new area of interest for IAVs researches. Even that, all the works exposed in this article have been tested only on simulations, the results show that Bayesian networks are a promising computational model to assist IAVs in the decision making process. Each of the articles represent a different situation that an IAV has to face. Taking the purposes as a starting point and the findings from the literature, the analysis is divided in three sections: (a) topics, (b) decision making framework and (c) Bayesian networks related aspects.

*(a) Topics*
Regardless of the means by which Autonomous Vehicles move, all of them in some moment of their performance need to make decisions. The articles are organized by means of transport. However, it can be noticed that the IAVs have to deal with similar aspects in the decision making process in a real world. Thus, the Bayesian networks have been used to model situations that involve the topics showed in

*Table 1*. The description explains the BN's specific functions. Looking at the publication year of the articles, it comes out that the topics are very new. Advances in vehicle automation have led to let them operate in the real world, without evaluating their impact first. Now the researchers are trying to model the behaviour of IAVs to integrate them into the peoples lives and the society complying with the established parameters.

Table 1. Topics related to IAVs modelled with Bayesian networks

| TOPIC | DESCRIPTION | REFERENCE |
|---|---|---|
| Landing site selection for scientific purposes | Selection of the best landing site in the space based on areas of scientific interest (*decision*). | [Serrano, 2006] |
| Situation assessment in a battle field | *Fusion* of information and analysis of variables to study the *opponents' reaction* (*prediction*). | [Cao et al., 2010] |
| Identify behaviour patterns | Detect unexpected behaviours or anomalies for threat assessment (anomaly *fusion*). | [Lane et al., 2010] |
| Dynamic threat assessment | Estimation and *prediction* of the states of the pop up threats. | [Gao et al., 2011] |
| Software health management | *Monitoring* of the functioning of the software and hardware and their *interaction* during system operation (*faults detection*). *Diagnostic* reasoning in order to identify the most likely root causes for the faults. | [Schumann et al., 2011] |
| Cooperation on search missions based on IAVs' capabilities | *Modelling* of unmanned autonomous vehicles' capabilities | [Guo et al., 2012] |
| Battle damage assessment | Damage assessment of the attack from IAV to enemies' ground airfield runway (*consequences' prognosis*). | [Chen-han and Jian, 2014] |
| Safety | IAV malfunctioning is evaluated taking into account transport regulations (*risk assessment*) | [Li et al., 2016] |
| Landing site selection in the event of an emergency | Selection of a safe landing site based on priorities (*decision*). | [Coombes et al., 2016] |
| Cyber-physical threats detection | *Diagnosis to determine* the origin of a threat (cyber attack or sensors attack). | [Bezemskij et al., 2017] |
| Level of autonomy | *Prediction* of level of autonomy for IAVs' supervision (*foresee* the mental workload of the human). | [Bazzano et al., 2017] |

| TOPIC | DESCRIPTION | REFERENCE |
|---|---|---|
| Human-IAV *relationship* | Evaluation of the elements for an adequate human-IAV *interaction* to ensure the mission success. *Trust* is an important variable for collaboration (*interactions*) | [Thieme and Utne, 2017] |
| Liability and responsibility | *Ditribution* of liability and responsibility in personal automotive transportation. | [Gopalswamy and Rathinam, 2018] |

*(b) Decision making framework for IAVs*

Since in the future the Intelligent Autonomous vehicles (IAVs) are going to move freely in the cities, countryside, sea, space or any area without a direct human control; the IAVs are going to have their own responsibilities and therefore have to make their own decisions. Cross a road, for example, means to make a decision. The car has to choose between stop to look for other ones crossing (cars or pedestrians) or to continue. Decision making for an Autonomous Vehicle is the process by which it makes a choice between different options or possible ways to solve different situations for its performance in different contexts. In other words, the decision making framework set the conditions to identify a problem and select an action course to solve it according to predefine goals. Based on the literature reviewed here, we state a decision making general framework of what it would mean for an autonomous vehicle to be able to make a decision. Making a decision would depend on several variables, such as, the ones shown in the decision making framework. *Table 2* shows the decision making framework for IAVs, it contains five modules and their respective variables related to the articles studied:

*Trigger point*. First at all, the autonomous vehicle has to recognised its current situation and to visualize its goals and priorities. Then the vehicle would look for ways to operate to reach its objectives. In a real world, the objectives do not come alone, they imply to take into account social, political, economical and government regulations. Depending on the situations, sometimes the vehicles can make decisions by their own and sometimes make group decisions. Then the vehicles would have to communicate to each other and to come to an agreement to define the way to operate collectively. The module trigger *point* does not have references because of is still under discussion, the moment when the vehicle has to make high level decisions involving free operation and interaction with humans and other vehicles is still under evaluation. Protocols need to be established. However, several advantages of IAVs' making decisions can be mentioned: their decisions are not affected by their state of mind or fatigue, they are the result of the evaluation of multiple options, they can manage more data than humans and predict consequences. All the articles presented give an overview of moments when the IAVs have to make decisions.

*Preparation*. The IAV has to know the kind of decisions it is able and allow to make (*level of autonomy*). When it is working in a group has to evaluate if can count on its partners (*trust*). To execute an action all its predefined functions need to be verified (*reliability on vehicle functions*). The context in which the decision is going to be made needs to be assess, circumstances or elements could be changing in time and space (situational awareness). Adequate forms to facilitate information exchange between IAVs and its partners, getting right information and in time is important (interaction). The IAVs also must show its reaction in a clear way. The activities for which the vehicle has to be ready are also determined. In summary, in the preparation process all the elements that defined the vehicle's role are established.

*Evaluation*. It consist of making a detailed study of every possible solutions that were generated for the problem, that is, looking at its advantages and disadvantages, individually with respect to the decision criteria, and one with respect to the other , assigning them a weighted value. The selection would be the one that adjust the most to the goals and criteria. Since we are talking about vehicles having a probabilistic model to quantitatively value their performance, the vehicles would be able to predict, according to their current state, in what degree they could meet their obligations. As well as, foresee the degree of posible damage that their decision could cause. Hence, it could study the probable consequences of its decisions and actions. The origin of the decision and the form the decision is being made is also measured.

*Execution*. Implement the decision made according to the chosen plan in order to evaluate whether the decision was successful or not. The action(s) should be monitor during operations for future evaluations. Until now, the decisions have been implemented on simulations and more simulations are needed to create statistics and different scenarios for possible situations on real life. Before execution; government, society and Autonomous Vehicles companies should come into an agreement to set up the manner IAVs will operate, this includes under what context they will performed (infrastructure conditions and participants). People should be informed about the vehicles skills and behaviour to know what to expect.

*Results evaluation*. Once the decision was executed, it is necessary to evaluate if the problem was solved or not, that is, if the decision is having the expected result or not.

In addition, we must be aware that the decisions that are taken continuously will have to be modified, due to the evolution of the situation or the appearance of new variables that affect it. All the modules of the decision making framework not necessarily have to be included in all the situations to make decisions. Maybe for the design and behaviour analysis of the IAV we could need all of them, but for performance in real time we have to select just some of them.

The studies exposed above take us to see that the Bayesian networks can let us define clear relationships between factors being part of the decision making framework, we can notice how they influence to each other, where the information and decision are coming from. Therefore, the causes and consequences of the decision can be evaluated. The interaction with other computational models is also possible, they can let us reason with information originated from other computational models or can share information to foresee events.

All the decision making modules have been study separately. We could think to have in the future a BN that integrates all the different modules. Consequently, we could count with a computational formalism to study how each module is affecting the decision making. The model could track what module of the decision making framework should be modified to improve the decision or tell us what module took the vehicle to a bad decision.

Table 2. Decision making framework for IAVs

| MODULES | VARIABLES | DEFINITION | REFERENCES |
|---|---|---|---|
| **Trigger point** | Goals, priorities, decision criteria | Determine the moment when the vehicle has to choose a course of action. Definition of its objectives, the relevance of the variables to reach them; how to assess the decision. | |
| | Information | Collect data from different sources | |
| **Preparation** | Level of autonomy | The systems' ability to make independent decisions. This depends on the type of operation to be carry out and the type of vehicle. | [Bazzano et al., 2017], [Thieme and Utne, 2017] |
| | Trust | How much the vehicles believe in the other part with which they are interacting according to the circumstances and their abilities. | [Thieme and Utne, 2017] |
| | Reliability on vehicle functions | The system's ability to perform its functions as required during the time of used. | [Bazzano et al., 2017], [Thieme and Utne, 2017] |
| | Situational awareness | Perception of the elements in the environment within a volume of time and space, the comprehension of their meaning and the projection of their status in the near future. We can have contexts of certainty, *risk or uncertainty*. | [Cao et al., 2010], [Li et al., 2016],[Thieme and Utne, 2017] |

| MODULES | VARIABLES | DEFINITION | REFERENCES |
|---|---|---|---|
| | Interaction | The way the vehicle collect information from the other vehicles, humans and its surroundings. The way it communicates with them and reacts. Includes: exchange of information, feedback from the system, interface design, time delay of transmission, etc. | [Thieme and Utne, 2017] |
| | Task definition | Work to accomplish according to goals or regulations. | [Guo et al., 2012] |
| **Evaluation** | Action plan | Analysis of possible ways to reach the goals. The plan could consider maximised benefits, less effort, less time, etc. | |
| | Responsibility | The fulfillment degree of obligations (tasks). | [Gopalswamy and Rathinam, 2018] |
| | Liability | Degree of vehicle fault for damage caused | [Gopalswamy and Rathinam, 2018] |
| | Causes | Why the decision has to be made? | [Schumann et al., 2011] |
| | Consequences | What could happened after a decision was made? Possible impact of the decision. | [Chen-han and Jian, 2014] |
| **Execution** | Actions | The vehicle has decided what to do and how to do it to reach the goals and it gets into action. It acts according to a plan or by simple reaction. | [Coombes et al., 2016], [Serrano, 2006] |
| **Results evaluation** | Feedback from the decision made | Do the goal was reach? Does the vehicle need more time to solve the problem? Was the decision made wrong? In this case the vehicle has to detect errors and to start again. | |

*(c) Bayesian networks related aspects*
When working with Bayesian networks, the first thing to think about is how to built one. They can be built based on data or experts knowledge. Prior to the construction of the network, it is important to have reliable and sufficient information from the area to be studied. It can also include information from expert systems. In the majority of the works studied here, they were built on expert knowledge. **[Thieme and Utne, 2017]**'s work attracts the attention, since they modelled the network in terms of document analysis (specialised literature).

The BNs creation for IAV could become a kind of qualitative (subjective) process. Nevertheless, after the operational framework of the IAV has been model in a BN, its behaviour and performance can be evaluated quantitatively. In some articles they used learning algorithms to find the relationships among variables. A problem is represented in a Bayesian network by selecting relevant variables from a domain, finding the relationships between variables and defining the probable estate for each variable. There are well-founded algorithms for that as those mentioned in section 3.

The way in which the authors of previous works present their approach, focuses mainly on the study of the problem and the definition of the structure of the network, although methods of learning, inference and software of Bayesian Networks are also mentioned. They show how various problems faced by Autonomous Vehicles can be studied from the principles of Bayesian networks. When different events occur, the variables can take different states and the probability that those states present themselves has to be estimated. Therefore, probabilities must be estimated that reflect the current situation of the problem and calculate probabilities that change according to the evolution of the situation (inference).

In some cases a preprocessing has been carried out or assumptions defined in order to reduce the number of nodes or to simplify the structure of the network. The reasoning process of a relatively small network could be done onboard for IAV. Networks of a bigger size could include variables to help in the design of IAV and predict its behaviour. In the literature many algorithms have been developed to build Bayesian Networks (learning) and algorithms so it is able to answer our questions (inference).

With regard to the shape of the network, the tree structure is the most employed, the algorithms developed for it give good results and are fast. However, when talking about networks integration for the whole decision making framework, they can become extremely big and computational time will increase. We could also get complicated networks for team performance evaluation. Besides the structure, the type of values that each variable can take can also complicate the execution time. Continuos variables will take more time for the algorithms to evaluate them. In many cases we just need to express the problems in qualitatively terms and we can discretised the variables. As showed in the literature.

## 5 Bayesian networks contributions to IAVs and challenges

As we can notice in *Table 1* there are several words in italics. This was made to remark the ways BN can support IAVs in their reasoning process to execute their tasks. In general we can say that BN let the vehicles to have a *model of the scenery* of a particular situation. That model could include all the variables of interest for the IAVs, the relationship between them, as well as, *fusion* information coming from different sources. The BN could help them to *monitor* their own state and the state of the other ones with whom are interacting. According to the structure and parameters of the net, the vehicles could be able to *predict* the behaviour of their partners in a team, other agents or objects involved in the situation. Since the model could represent a whole view of the situation, better decisions could be made. The vehicles could reason the causes and consequences of such decisions. Thus, IAVs would be aware of their actions and generate detail behaviour sequences.

In the *Figure 1*, we introduced an operation and interaction framework of IAVs considering intrinsic and extrinsic factors. The articles discussed here give examples of how BNs could integrate those factors.

The graphic and probabilistic nature of Bayesian networks gives them the following strengths applicable in the performance of IAVs:

- Deal with uncertainty in random environments (pop up objects)
- Able to work with missing data (sensor broken)
- Detect abnormal behaviour (patterns)
- Identify and classify the human intention (from the other drivers and pedestrians)
- Reason (inference, manipulate network information)
- Interpret (understand behaviour)
- Model complex environments (characteristics of the agents or events and their relationship between them and the scenery)
- Model activities consisting of a group of events or interactive activities of a group of agents
- Model human-IAV cooperation and interaction, software-hardware's IAV interaction
- Monitor the performance of the autonomous vehicle
- Fusion different sources of information (sensors, databases, other networks, reasoning tools, information about different situations like on a battlefield)
- Manage changes of the variables in the time (Dynamic Bayesian networks)
- Form of giving a quantitative and qualitative response to decision making
- Diagnose which parts generated the problem and in what proportion (calculation of faults in accidents)
- Predict consequences if variables probabilities are updated given evidence (information) we have about some variables

Thinking of an IAVs architecture, BNs can be seen as a tool to assist the kind of conceptual modules such as those shown in *Table 3*.

Table 3 Conceptual modules of an IAVs' architecture assisted by BN

| Module | How BNs assist |
|---|---|
| Risk assessment | Predict the cost (consequences) of their actions or misfunctioning. |
| Fault detection and diagnosis | Identify which one of their components are failing and why (origin of the fail). |
| Prediction | Anticipate their actions to the events changing in the environment (deal with uncertainty). |
| Mission planning and goal selection | Model and analysis of complex situations. |
| Human-vehicle interaction and social cooperation | Model human and IAVs' capabilities and behaviours and the relationships between them. Supervision of the performance of the Autonomous Vehicles. |
| Situation awareness | Fusion of information from different sources to detect abnormal behaviour (behaviour patterns) and interpret it. |
| Safety requirement | Include safety standards (government, organisations) |
| Decision making | Evaluate the execution of selected actions taking into account level of autonomy, trust, responsibility, liability, causes and consequences. |

In a real world IAVs could start to perform with a predefine BN. However, they would face a changing world. So they are challenged to adapt their knowledge and observations to new situations. The construction of a net dynamically, according to the evolution of the situation is still under research. This involves to structure the knowledge acquisition to built the net. A way to extract the most relevant variables should be also defined. Besides the network construction, the interaction between BNs and other computational models should be researched to develop systems working together for the operation and interaction framework (*Figure 1*) of IAV, considering intrinsic and extrinsic factors. Another important aspect to cover is to state a process to validate the recent acquired data and the built of new networks.

# 6 Towards safe, collaborative IAVs

Human safety is the priority for means of transport. Since Intelligent Autonomous Vehicles would interact with humans, this should be also their priority. We require that IAVs have a predictable behaviour and that their actions are in accordance with the regulations. It is especially important that humans have confidence in them so that they have the will to collaborate with them. There are missions, such as rescue missions, that require of a team participating. In a team each member performs better when he knows the way of working of other members and then is able to adapt to their way of working or humans behaviours. So trust is an important factor for mission success. With confidence come the levels of autonomy, the more certain we are of the vehicle's abilities, the more tasks we can delegate to it. The autonomous vehicle itself should be able to recognize its own capabilities and together with its level of autonomy make decisions. In addition to its capabilities and level of autonomy, a well structured decision process should be followed, we can see an analogy with the decision making process of pilots. Pilots have a handbook of aeronautical knowledge in which aeronautical decision making is profiled (this handbook is published by the United States Department of Transportation, http://www.flyinhighokc.com/pilotshandbook/pages/0-3.html). In the same way, decision making protocols could be defined for the different type of Intelligent Autonomous Vehicles.

Each one of the modules described in *Table 3* contribute in some degree to reach safety in Intelligent Autonomous Vehicles. They constitute a bridge between Intelligent Autonomous Vehicle technical performance and its relationship with its environment and humans.

# 7 Conclusions

The works in this survey have been presented in chronological order and grouped according to the means by which they move. Cases of interest to the international community that affect the activities of people's daily lives were exposed. The test cases were described, specifically in what part of the problem Bayesian networks participate, how they have been built, size of the networks, types of variables, learning and inference algorithms they use and software to work with them. The type of variables studied in the different problems show the kind of information can be managed in the network's nodes (qualitative, quantitative, discrete, continuous). The most common network structured observed is a tree. Sometimes independence assumptions have been made between variables relationships to simplify network structure. The results presented in the articles are based on simulations, so remains for the future to apply BNs on real data. This poses a challenge in the data acquisition process. Tests are missing in real situations, BNs construction requires knowledge of the study area and large amounts of data to correctly model them. The integration of the IAVs into the society means BNs should also reflect ways of thinking, opinions, political and economical situations, laws and social behaviours (extrinsic factors). The automatic creation of the representative net of a study problem requires a structured form for selecting relevant parameters. In a real world IAVs always are going to find changing situations, so they can not decide how to act just with a simple yes or not. BNs are a good tool for IAVs to evaluate their possible actions and to make decisions in an uncertain environment. Figuring an IAVs' architecture, BNs could help the top layers in which we can have modules for self diagnosis (recognize what IAVs is able to do or its own failures), receiving high level commands and generate sequences of actions (mission planning), supervision to monitor the performance of a team of IAVs vehicles and coordinating operations (decision making).

As a result of this survey three tables were created. The first table gives an overview of the applications in the decision making for IAVs, has some very interesting and topical issues. It is not so much comparative, rather demonstrative since the applications found cover very varied topics and each one of them contributes something to the decision making framework. It is a table that give us a starting point to know what elements to think about in order to structure decision making. The second one is a table that according to the literature and analysis done in this survey is how you could see a general decision making framework of the Autonomous Vehicles. It includes the modules and variables of the framework. In addition, this two tables put together what has been done so far related to IAVs' decision making, they give you a clear idea of where the research is going and since when work has been done on the decision making, where the specific functions of Bayesian networks are and we can also notice non deterministic situations IAVs have to face. The third table suggest some modules of an IAVs' architecture that could be assisted by BNs and describe how BNs help them.

The most frequently use of Bayesian networks applied to IAVs has been in high level decision making. Currently, great care has been taken to allow Autonomous Vehicles to make decisions that involve a risk to humans. With the advancement of technologies in this field, IAVs will increasingly make more decisions on their own, without human supervision. The decisions could be made individually or collectively and new models are needed to represent and study the decision making process. Although the results presented are based on simulations, they show that BNs could help the designers of IAVs to realise a sensitivity analysis to get a deep view of the autonomous system and find the key factors of IAVs security. The designers could get a feedback in advanced predicting their behaviour.

In addition BNs let us develop models to provide a transformation of high level contextual information to lower level information that is suitable and understandable for the autonomous systems.

For the decision making framework of a specific IAV, namely, taking into account the means where is moving, protocols should be established to create a framework for IAVs to make decisions under intrinsic and extrinsic factors. For example, a handbook could be developed for Aerial Unmanned Autonomous Vehicles in the same way that a handbook for pilots exists, which includes aeronautical decision making.

31 de 34Michael Milford, Peter Corke, "*The Limits and Potentials of Deep Learning for Robotics,*" arXiv:1804.06557v1 [cs.RO] 18 Apr 2018.

**[Gopalswamy and Rathinam, 2018]** Swaminathan Gopalswamy and Sivakumar Rathinam, "*Infrastructured Enabled Autonomy: A Distributed Intelligence Architecture for Autonomous Vehicles,*" arXiV:1802.04112v1 [cs.CY] 5 Feb 2018.

**[Pearl, 1985]** Judea Pearl "*Bayesian Networks: A Model of Self-Activated Memory for Evidential Reasoning,*" (UCLA Technical Report CSD-850017) in *Proceedings of the 7th Conference of the Cognitive Science Society, University of California, Irvine, CA*. : 329–334. 1985.

**[Kochenderfer, 2015]** Mykel J. Kochenderfer, "Decision Making Under Uncertainty, Theory and Application," The MIT Press, 2015.

**[Thieme and Utne, 2017]** Christoph Alexander Thieme and Ingrid Bouwer Utne, "A Risk Model for Autonomous Marine Systems and Operation Focusing on Human-Autonomy Collaboration," Proc IMechE Part O: J Risk and Reliability, Vol. 231(4) 446-464, 2017.

**[Cao et al., 2010]** Lu Cao, An Zhang and Qiang Wang, "Research on Situation Assessment of UCAV Based on Dynamic Bayesian Networks in Complex Environment," Springer-Verlag Berlin HeidelBerg, 2010.

**[UKAutodrive, 2018]** UKAutodrive project, http://www.ukautodrive.com/the-uk-autodrive-project/ (accessed 4, June 2018).

**[DMV, California, 2018]** California Department of Motor Vehicles, https://www.dmv.ca.gov/portal/dmv/detail/vr/autonomous/testing?lang=en (accessed 4 June, 2018).

**[macReports, 2018]** Apple macReports, https://macreports.com/apple-again-grows-its-california-self-driving-fleet/ (accessed 4 June, 2018).

**[Kongsberg, 2016]** https://www.km.kongsberg.com/ks/web/nokbg0397.nsf/AllWeb/FA364D4791C19536C12580F70024BAC4/$file/Flyer-test-bed-autonomous-shipping.pdf?OpenElement G(accessed 4 June, 2018).

**[Kongsbers, 2017]** https://www.km.kongsberg.com/ks/web/nokbg0238.nsf/AllWeb/FA9523C22BEB1F0DC12581EF0033AA86?OpenDocument (accessed 4 June, 2018).

**[EASA, 2017]** EASA–FAA International Safety Conference, 14-16 June, Brussels, 2017.

**[Starek et al., 2016]** Joseph A. Starek, Behçet Açikmeşe, Issa A. Nesnas, Marco Pavone, "Spacecraft Autonomy Challenges for Next-Generation Space Missions. " In: Feron E. (eds.) Advances in Control System Technology for Aerospace

34 de 34**[Doshi et al., 2002]** Prashant Doshi, Lloyd Greenwald and John Clarke, "Towards Effective Structure Learning for Large Bayesian Networks", American Association for Artificial Intelligence, 2002.

**[Heckerman, 1996]** David Heckerman, "A Tutorial on Learning with Bayesian Networks", technical report MSR-TR-95-06, Microsoft Research, Advanced Technology Division, 1996.

**[Jensen and Nielsen, 2007]** Finn V. Jensen and Thomas D. Nielsen, "*Belief Updating in Bayesian Networks*," Chapter in Bayesian Networks and Decision Graphs. Information Sciences and Statistics. Springer, New York, NY, 2007.

**[Russell and Norvig, 2009]** Stuart Russell and Peter Norvig, "*Artificial Intelligence a Modern Approach*," third edition, Prentice Hall, 2009. ISBN 0-13-604259-7

**[Kwisthout, 2011]** Johan Kwisthout, "*Most Probable Explanations in Bayesian Networks: Complexity and tractability*," International Journal of Approximate Reasoning, 52 (2011), 1452-1469, 2011.

**[Van der Gaag et at., 2007]** Linda C. Van der Gaag, Silja Renooij and Veerle Coupé, "Sensitivity Analysis of Probabilistic Networks", **Chapter** *in* Studies in Fuzziness and Soft Computing, December 2007.

**[Zuk et al., 2006]** Or Zuk, Shiri Margel and Eytan Domany, "*On the Number of Samples Needed to Learn the Correct Structure of a Bayesian Network*", in Proceedings of the Twenty-Second Conference on Uncertainty in Artificial Intelligence (UAI2006).

**[Friedman et al., 1999]** Nir Friedman, Moises Goldszmidt and Abraham Weyner, "*Data Analysis with Bayesian Networks: A Bootstrap Approach*", in Proceedings of the Fifteenth Conference on Uncertainty in Artificial Intelligence (UAI1999).

**[Schietekat et al., 2016]** Sunelle Schietekat, Alta de Waal and Kevin G. Gopaul, "*Validation & Verification of a Bayesian Network Model for Aircraft Vulnerability*", 12th INCOSE SA Systems Engineering Conference, ISBN 978-0-620-72719-8, 2016.

**[Ron, 1995]** Ron Kohavi, "A Study of Cross-Validation and Bootstrap for Accuracy Estimation and Model Selection", in the International Joint Conference on Artificial Intelligence (IJCAI), 1995.